\documentclass[twoside]{article}

\usepackage[accepted]{aistats2024}
\usepackage{graphicx}
\usepackage{latexsym}
\usepackage{shortcuts} 
\usepackage{adjustbox}
\usepackage{booktabs}  

\usepackage{microtype}
\usepackage{booktabs} 
\usepackage[hidelinks]{hyperref}
\usepackage{url}
\usepackage{float}
\usepackage{subcaption}
\usepackage{tikz}
\usepackage{amssymb}
\usepackage{amsmath}
\usepackage{multirow}

\usetikzlibrary{decorations.pathreplacing}

\usepackage{amsmath}
\usepackage{amssymb}
\usepackage{mathtools}

\begin{document}

%

%
\runningauthor{F. Carton, R. Louiset, P. Gori}

\twocolumn[

\aistatstitle{Double InfoGAN for Contrastive Analysis}

\aistatsauthor{Florence Carton$^1$ \And Robin Louiset$^{1,2}$ \And  Pietro Gori$^1$ }

\aistatsaddress{$^1$LTCI, Télécom Paris, IPParis, France \And $^2$NeuroSpin, CEA, Universite Paris-Saclay, France } ]

\begin{abstract}
Contrastive Analysis (CA) deals with the discovery of what is common and what is distinctive of a target domain compared to a background one. This is of great interest in many applications, such as medical imaging. Current state-of-the-art (SOTA) methods are latent variable models based on VAE (CA-VAEs). However, they all either ignore important constraints or they don't enforce fundamental assumptions. This may lead to sub-optimal solutions where distinctive factors are mistaken for common ones (or viceversa). Furthermore, the generated images have a rather poor quality, typical of VAEs, decreasing their interpretability and usefulness. Here, we propose Double InfoGAN, the first GAN based method for CA that leverages the high-quality synthesis of GAN and the separation power of InfoGAN. Experimental results on four visual datasets,
from simple synthetic examples to complex medical images, show that the proposed
method outperforms SOTA CA-VAEs in terms of latent
separation and image quality. Datasets and code are available online$^{1}$.
\end{abstract}

\section{Introduction}
\label{intro}
Learning disentangled generative factors in an unsupervised way has gathered much attention lately since it is of interest in many domains, such as 
medical imaging. Most approaches look for 
factors that capture distinct, noticeable and semantically meaningful variations in \textit{one} dataset (e.g., presence of hat or glasses in CelebA). Authors usually propose well adapted regularizations, which may promote, for instance, "uncorrelatedness" (FactorVAE \cite{kim_disentangling_2018}) or "informativeness" (InfoGAN \cite{chen_infogan_2016}).\\
In this paper, we focus on a related but \textit{different} problem, that has been named Contrastive Analysis (CA) \cite{zou_contrastive_2013,abid_exploring_2018,weinberger_moment_2022}. We wish to discover in an unsupervised way what is added or modified on a target dataset compared to a control (or background) dataset, as well as what is common between the two domains. For example, in medical imaging, one would like to discover the salient variations characterizing a pathology that are only present in a population of patients and not in a population of healthy controls. Both the target (patients) and the background (healthy) datasets are supposed to share uninteresting (healthy) variations. The goal is thus to \textit{identify} and \textit{separate} the generative factors common to both populations from the  ones distinctive (i.e., specific) only of the target dataset.\\
The most recent CA methods are based on the Variational AutoEncoders (VAE) \cite{kingma_auto-encoding_2014} model and they are called Contrastive VAE (CA-VAE). These methods assume that samples from the target dataset are generated using two sets of latent factors, common $\bz$ and salient $\bs$, whereas samples from the control dataset are generated using only the common $\bz$ factors. The salient factors $\bs$ should therefore model the specific patterns of variations of the target dataset. All these methods share the same general mathematical formulation, which derives from the standard VAE. However, they all either ignore a term of the proposed loss (e.g., KL loss in \cite{abid_contrastive_2019,ruiz_learning_2019}) or they don't enforce important assumptions (e.g., independence between $\bz$ and $\bs$ in \cite{weinberger_moment_2022}), which may lead to sub-optimal solutions where salient factors are mistaken for common ones (or viceversa).
Furthermore, they all share a typical downside of VAEs: a blurry and poor quality image generation.\\
For these reasons, we propose \textit{Double InfoGAN}: a novel Contrastive method which leverages the high-quality synthesis of Generative Adversarial Networks (GANs) \cite{Goodfellow2014} and the separation power of InfoGAN \cite{chen_infogan_2016}. To the best of our knowledge, this is the first GAN based method proposed in the context of Contrastive Analysis. The main contributions of this paper are:

\noindent $\bullet$ The first GAN based method for Contrastive Analysis (CA) which allows high-quality synthesis.\\
$\bullet$ A new regularization term for CA, inspired by InfoGAN.\\
$\bullet$ Two new losses for an accurate separation and estimate of the common and salient generative factors.\\
$\bullet$ Extensive experimental results on four visual datasets, from synthetic to complex ones, show that the proposed method outperforms SOTA CA-VAE methods in terms of latent separation and image quality. Datasets and code are available online.\footnote{\url{https://github.com/Florence-C/Double_InfoGAN.git}}

\begin{figure}[t]
    \centering
    \includegraphics[width=\linewidth]{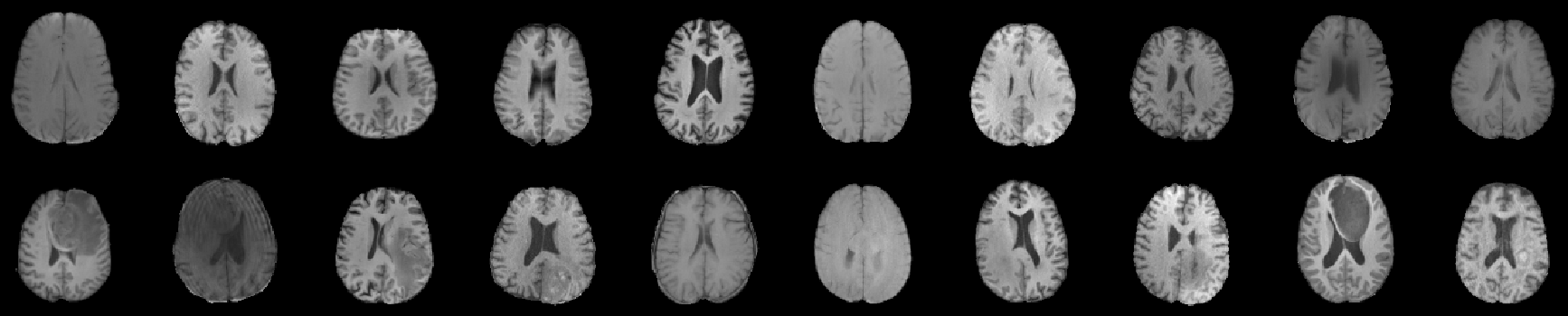}
    \includegraphics[width=\linewidth]{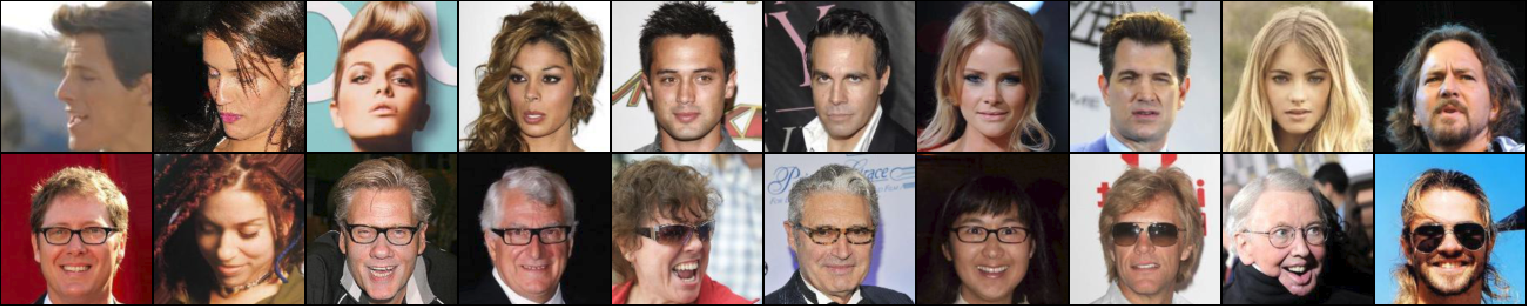}
    \caption{Two examples of datasets for Contrastive Analysis. \textbf{First figure}: Brats dataset \cite{menze2014multimodal}. Top: MRI images of healthy brains (control dataset). Bottom: MRI images of brains with tumor (target dataset). \textbf{Second figure}: CelebA dataset. Top: control dataset with regular faces (no smile, no glasses). Bottom: target dataset that contains smiling faces with glasses.}
    \label{fig:brats}
\end{figure}

\section{Related Work}
Separating common from distinctive latent representations has become an active research area in several fields, such as domain adaptation (DA) \cite{ganin_domain-adversarial_2017,hoffman_cycada_2018} and image-to-image translation (IMI) \cite{zhu_unpaired_2017,isola_image--image_2017,liu_unsupervised_2017,ferrari_diverse_2018,huang_multimodal_2018}.\\ 
\textbf{DA} seeks to transfer a classifier from a source domain, with many labelled samples, to a different target domain, which has few or no labelled data. As shown in \cite{ganin_domain-adversarial_2017}, an effective classifier should use shared features that cannot discriminate between the two domains. The goal of \textbf{IMI} is instead to estimate a transformation that maps images from the source domain to the target one by disentangling and controlling high-level visual attributes (style, gender, objects) \cite{ferrari_diverse_2018}. The main difference between these methods and the proposed one is the objective. Our goal is to statistically analyze two domains (e.g., healthy and patients) looking for latent representations that generate the background (e.g., healthy) and target (e.g., pathological) content. We do not seek to transfer a classifier or to map an image to a different distribution. We wish, for instance, to generate new images and not only to translate them to another domain. Another important difference is that we do not want to encode only a particular distinctive attribute (e.g., style \cite{ma_exemplar_2019}, gender) but \textit{all} distinctive variations of the target domain with respect to the background one. Furthermore, we do not plan to use a weight sharing constraint \cite{ferrari_diverse_2018,liu_unsupervised_2017}, or other architectural constraints, which assume that the main differences are, for instance, only in the low-level features (color, texture, edges, etc.).\\ 
Our work is also close to \textbf{unsupervised anomaly detection }~\cite{guillon_detection_2021,Baur2021,pang_deep_2022,vetil_learning_2022}, which is usually composed of two steps. First, the distribution of the background (control) domain is learned, using deep generative models. Then, or at the same time, a discriminator is optimized to detect the target (anomalous) samples. By looking at the reconstruction errors \cite{guillon_detection_2021}, attention scores \cite{vedaldi_attention_2020}, visual saliency \cite{kimura_adversarial_2020} or other features, one can understand which are the salient patterns of the target (anomalous) domain. Even if this strategy can be highly interpretable, the goal is to spot an anomalous sample and not to model the latent factors that generate the anomalous patterns.\\
 Another class of methods, mainly used in the fields of data integration and data fusion, are the \textbf{projection based latent variables approaches}, such as 2B-PLS, 02PLS, DISCO-SCA, GSVD, JIVE \cite{feng_angle-based_2018,rohlf_use_2000,deun_disco-sca_2012,yu_jive_2017,trygg_o2-pls_2002,smilde_common_2017}. Contrary to these methods, we do not use only linear transformations, but we leverage the capacity of deep learning to estimate non-linear mappings.\\
 In parallel, research on \textbf{disentanglement} has been developed, making it possible to modify a single and semantically meaningful pattern of the image (e.g., person's smile, gender), by varying only one component of the latent representation \cite{kim_disentangling_2018}. As shown in \cite{locatello_challenging_2019}, the unsupervised learning of disentangled representations is theoretically impossible from i.i.d. samples without inductive biases \cite{higgins_-vae_2017,chen_infogan_2016}, weak labels \cite{shu_weakly_2020,locatello_weakly-supervised_2020}, or supervision \cite{Lample2017a,Choi2018,He2019,Shi2021,joy_capturing_2021}. These methods have all focused on the latent generative factors of a \textit{single} dataset, and their goal is thus different from ours.\\
With a different perspective, methods stemming from the recent \textbf{Contrastive Analysis (CA)} setting \cite{zou_contrastive_2013,abid_exploring_2018,tu_capturing_2021,ruiz_learning_2019,zou_joint_2022,abid_contrastive_2019,choudhuri_towards_2019,severson_unsupervised_2019,weinberger_moment_2022} mainly use variational autoencoders (VAE) to model latent variations only present among target samples and not in the background dataset. Similarly, in \cite{Benaim2019}, authors used standard autoencoders to estimate common latent patterns between two domains as well as patterns unique to each domain. Being based on auto-encoders, this method cannot sample in the latent space (i.e., no new image generation) and its goal is to map sample images from one domain to the other, as in IMI. Another related method is NestedVAE~\cite{vowels_nestedvae_2020}, whose goal is bias reduction by estimating common factors between visual domains using \textit{paired} data. Here, we wish to use unpaired datasets.\\
Lastly, CA is different from \textbf{style vs. content separation} and \textbf{style transfer}. In particular, in recent works \cite{kazemi_style_2019, von_kugelgen_self-supervised_2021}, content usually refers to the invariant generative factors across samples and views (i.e., transformations/augmentations of a sample), while style refers to the varying factors. Content and style thus depend on the chosen semantic-invariant transformations, and they are defined for a single dataset. In CA, we do not necessarily need transformations or views, and we jointly analyze two different datasets.

\section{Background}
\textbf{InfoGAN} 
In \cite{chen_infogan_2016}, differently from standard GAN \cite{Goodfellow2014}, authors propose a new method, called InfoGAN, where they decompose the input noise vector of GANs into two parts: 1) $\bz$, which is considered as a nuisance and incompressible noise and 2) $\bc$, which should model the salient semantic features of the data distribution. The generator of this new model, $G(\bz,\bc)$, takes as input both $\bz$ and $\bc$ to generate samples $\bx$. As shown in \cite{chen_infogan_2016}, without regularisation, the generator $G$ may ignore the additional code $\bc$ or find a trivial (and useless) solution. To this end, authors propose to regularize the estimate of $G$ by maximizing the mutual information $I(\bc; \bx)$ between $\bc$ and $\bx \sim G(\bz,\bc)$. Maximum $I$ is obtained when $\bc$ and $\bx$ are completely dependent and one becomes completely redundant with the knowledge of the other. This should increase the informativeness of $\bc$, namely all salient semantic information should be in $\bc$ and not in $\bz$, which should only account for additional randomness (i.e.,  noise). Authors propose to maximize a lower bound of $I(\bc; \bx)$ by defining an auxiliary distribution $Q(\bc | \bx)$, parameterized as a neural network, to approximate $P(\bc|\bx)$:
\begin{equation}
 I(\bc;\bx) \geq \bbE_{\bz \sim P(\bz), \bc \sim P(\bc),\bx \sim P(\bx |\bc, \bz)} \log(Q(\bc|\bx)) + H(\bc)
\label{eq:InfoGAN-I}
\end{equation}
More mathematical details in the Supplementary.

\textbf{Contrastive VAE (CA-VAE)} 
In this section, we present the CA-VAE models \cite{choudhuri_towards_2019,severson_unsupervised_2019,abid_contrastive_2019,ruiz_learning_2019,zou_joint_2022,weinberger_moment_2022}. Let $X=\{\bx_i\}$ and $Y=\{\by_j\}$ be the background (or control) and target data-sets of images respectively. Both $\{\bx_i\}$ and $\{\by_j\}$ are assumed to be i.i.d. from two different  and unknown distributions ($P(\bx)$ and $P(\by)$) that depend on a pair of latent variables ($\bz$, $\bs$). Here, $\bs$ is assumed to capture the salient generative factors proper only to $Y$ whereas $\bz$ should describe the common generative factors between $X$ and $Y$. The generative models (i.e. same decoder with parameters $\theta$) are: $\bx_i \sim P_\theta (\bx | \bz_i, \bs_i=s')$ and $\by_j \sim P_\theta (\by_j | \bz_j, \bs_j)$, where the salient factors $\bs_i$ of $X$ are fixed to a constant value $s'$
(e.g., $s'=0$), thus enforcing $\bz$ to fully encode alone $X$. 
The conditional posterior distributions are approximated using another neural network (i.e. encoder with parameters $\phi$) shared between $X$ and $Y$, $Q_\phi(\bz_i,\bs_i|\bx_i)$ and $Q_\phi(\bz_j,\bs_j|\by_j)$, which are usually assumed to be conditional independent: 
$Q_\phi(\bz,\bs|\cdot)=Q_\phi(\bz|\cdot)Q_\phi(\bs|\cdot)$. The latent generative factors ($\bz$, $\bs$) are also usually assumed to be independent (i.e., $P_{\cdot}(\bz,\bs)=P_{\cdot}(\bz) P_{\cdot}(\bs)$). The common factor $\bz$ should follow the same prior distribution in $X$ and $Y$
(e.g., $P_x(\bz)=P_y(\bz)=\cN(\bz;0,\cI)$). 
The salient factor $\bs$ follows instead a different prior distribution between $X$ and $Y$, such as 
$P_y(\bs)=\cN(\bs;0,\cI)$ and $P_x(\bs)=\delta(\bs=s')$, the Dirac distribution centered at $s'$.
Based on this generative latent variable model, one can derive a lower bound of the marginal log likelihood: 
\begin{equation}
\begin{split}
\log P(\bx) \geq & \bbE_{Q_\phi(\bz|\bx)Q_\phi(\bs|\bx)}\log P_\theta(\bx | \bz,\bs)- \\ &KL(Q_\phi(\bz|\bx) || p_\bx(z))- KL(Q_\phi(\bs|\bx) || p_\bx(s))
\end{split}
\end{equation}
and similarly for $\log P(\by)$. All existing CA-VAE methods share this mathematical framework. They mainly differ for optimization or architectural choices and new added losses. However, none of these methods explicitly enforces the independence between common and salient latent factors\footnote{\cite{abid_contrastive_2019} proposed to minimize the total correlation (TC) between $q_{\phi_z,\phi_s}(z, s | x)$ and $q_{\phi_z}(z|x) q_{\phi_s}(s|x)$ via the density-ratio trick \cite{kim_disentangling_2018}, but their implementation is inaccurate since they don't use an independent optimizer.} and most of them ignore the KL divergence term about $p_\by(s)$ (except \cite{choudhuri_towards_2019} and \cite{weinberger_moment_2022}), thus allowing a possible information leakage between salient and common factors, as discussed in \cite{weinberger_moment_2022}. Furthermore, the quality of the generated images is rather poor.


\section{Proposed method - Double InfoGAN}
\label{method}

\begin{figure}[t]
\centering
\includegraphics[width=0.5\textwidth]{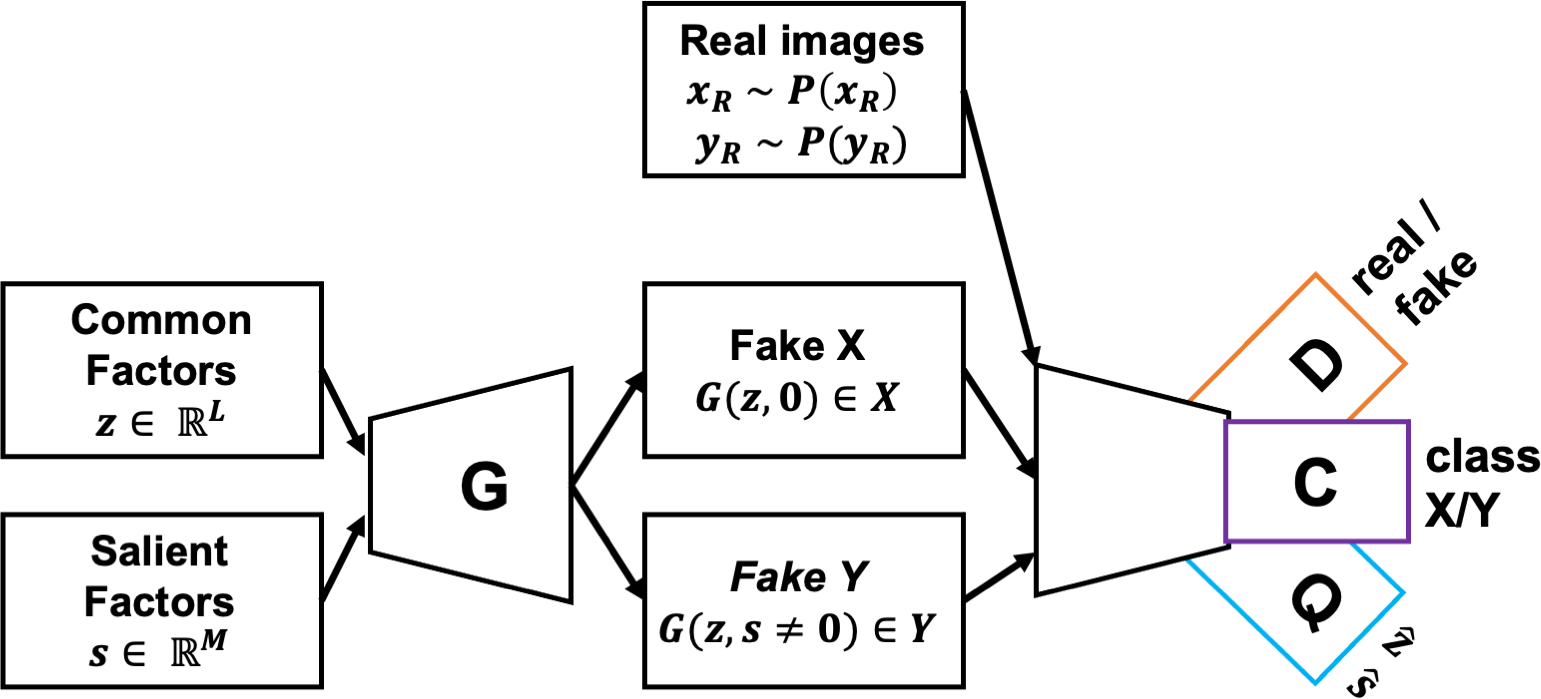}
  \label{fig:method}
    \caption{\textbf{Double InfoGAN}. Our model takes two inputs: $\bz$ (common factors) and $\bs$ (salient factors). The generator $G$ produces fake images that, together with the real images, are passed to a discriminator and encoder. The discriminator has two modules: $D$ for detecting real from fake images, and $C$ for classyfing images in the correct domain (i.e., $X$ or $Y$). The encoder $Q$ has two modules, $Q_z$ and $Q_s$, to reconstruct the latent factors $(\bz,\bs)$. $D$, $C$ and $Q$ share all layers but the last one.}
\end{figure}  

\textbf{Model} In Double InfoGAN, we use a generative model similar to the one proposed in CA-VAE but within the framework of InfoGAN. 
We suppose that the background images $\{\bx_i\} \overset{\text{iid}}{\sim} P(\bx)$ and the target images $\{\by_j\} \overset{\text{iid}}{\sim} P(\by)$, where $P(\bx)$ and $P(\by)$ are unknown and depend on a pair of latent variables ($\bz \in \bbR^{L}$, $\bs \in \bbR^{M}$). Differently from InfoGAN, and similarly to CA-VAE, $\bz$ should now capture the generative factors common to both $X$ and $Y$ whereas $\bs$ the salient factors proper only to $Y$. As in GAN \cite{Goodfellow2014}, we introduce a generator $G$ and a discriminator. The generator $G$ should generate samples that are indistinguishable from the true ones, whereas the discriminator is divided into two modules. The first (and standard) one $D$ is trained to discriminate between fake and real samples. The second module $C$ is trained to correctly classify real samples (i.e., $X$ or $Y$). 
As in InfoGAN, we also use one encoder, divided into two modules, $Q_z$ and $Q_s$, to reconstruct the latent factors $\bz$ and $\bs$. The discriminator, $D$ and $C$, and the encoder, $Q_z$ and $Q_s$, are parametrized as neural networks, that share all layers but the output one.\\ 
Let $\bx=G(\bz,\bs=s')$ and $\by=G(\bz,\bs)$ be the generated samples. We suppose, and force it in practice, that the latent variables $\bz=\{z_1,...,z_L \}$ and $\bs=\{s_1, ..., s_M \}$ are independent and follow a factorized distribution: $P(\bz)=\prod_{i=1}^L P(c_z)$ and $P(\bs)=\prod_{j=1}^M P(s_j)$, for $X$ and $Y$. The total cost function is:
\begin{equation}
\begin{split}
    \underset{G,Q_z,Q_sC}{min} \underset{D}{max} \quad w_{Adv}  \mathcal{L}_{Adv}(G,D)  + w_{Class} \mathcal{L}_{cl}(G,C) -\\ 
    w_{Info}  \mathcal{L}_{Info}(G,Q_z,Q_s) + w_{Im} \mathcal{L}_{Im}(G,Q_z,Q_s) 
\end{split}
\label{eq:Total-Loss}
\end{equation}
\noindent In the following, we will describe each term.

\textbf{Adversarial GAN Loss} \quad As in \cite{Goodfellow2014}, $G$ and $D$ are trained together in a \textit{min-max} game using the original nonsaturating GAN (NSGAN) formulation: 
\vspace{-0.1cm}
\begin{equation}
\begin{split}
    &\mathcal{L}_{Adv}(D,G) = w_{bg}  \Bigl( - \mathbb{E}_{\bx_R\sim P(\bx_R)}\bigl[\log(D(\bx_R)\bigr] -\\ 
    &\mathbb{E}_{z \sim P_x(\bz)}\bigl[\log(1-(D(G(\bz,0))))\bigr] \Bigr) +
    w_t \Bigl( - \mathbb{E}_{\by_R\sim P(\by_R)}\\ 
    &\bigl[\log(D(\by_R)\bigr] - \mathbb{E}_{\bz,\bs \sim P_y(\bz,\bs)}\bigl[\log(1-(D(G(\bz,\bs))))\bigr]\Bigr)
\end{split}
\end{equation}
\noindent where $D(I)$ indicates the probability that $I$ is real or fake and $\bx_R \sim P(\bx_R)$ and $\by_R \sim P(\by_R)$ are real images. 
Furthermore, we choose the same factorized prior distribution $P(\bz)$ for both $X$ and $Y$ (i.e., $P_x(\bz)=P_y(\bz)=P(\bz)$), namely a Gaussian $\cN(0,1)$. We also tested a uniform distribution $\cU_{[-1,1]}$ but the results were slightly worse. Instead, about $P(\bs)$, it should be different between $X$ and $Y$. We use a Dirac delta distribution centered at 0 for $X$ (i.e., $P_x(\bs) =\delta(\bs = 0)$) and we have tested several distributions for $P_y(\bs)$. Depending on the data and related assumptions, one could use, for instance, a factorized uniform distribution, $\cU_{(0,1]}$, or a factorized Gaussian $\cN(0,1)$ (ignoring the samples equal to 0). In our experiments, results were slightly better when using $\cN(0,1)$.

\textbf{Class Loss} \quad To make sure that generated images belong to the correct class, we propose to add a second discriminator module $C$. It is trained on real images to predict the correct class: $X$ or $Y$. At the same time, $G$ is trained to produce images correctly classified by $C$. We (arbitrarily) assign 0 (resp. 1) for class $X$ (resp. $Y$) and use the binary cross entropy ($\cB$). The loss is:
\vspace{-0.3cm}
\begin{equation}
\begin{split}
\mathcal{L}_{cl}(C) =  &\mathbb{E}_{\bx_R \sim P(\bx_R)}\bigl[\cB(C(\bx_R), 0)\bigr]  \\
& +\mathbb{E}_{\by_R \sim P(\by_R)}\bigl[\cB(C(\by_R), 1)\bigr]\\
\mathcal{L}_{cl}(G) = &\mathbb{E}_{\bz\sim P_x(\bz)}[\cB(C(G(\bz,0)), 0)] \\
& +\mathbb{E}_{\bz,\bs \sim P_y(\bz,\bs)}[\cB(C(G(\bz,\bs)), 1)]
\end{split}
\label{eq:loss-class}
\end{equation}

\textbf{Info Loss} \quad Similarly to InfoGAN, we propose two regularization terms based on mutual information, $I((\bz,\bs);\by)$ and $I((\bz,\bs=s');\bx)$, to encourage informative latent codes. However, in our case, these two terms are \textit{not} added to disentangle between informative and nuisance generative factors, but to enforce \textit{the separation} between common and salient factors. Indeed, the maximization of these two regularity terms should enforce $\bz$ to fully encode $X$ and at the same time to be informative for the generation of $Y$. In parallel, $\bs$ should only encode distinctive semantic information of $Y$.
Please note that the inclusion of two other nuisance factors, similarly to InfoGAN, describing the incompressible noise of $X$ and $Y$, would make the analysis more complex (i.e., additional regularity terms) since they should not model the common nor the salient generative factors.\\ 
Since $\bz$ and $\bs$ are independent \textit{by construction}, the mutual information $I((\bz,\bs); \cdot)$ can be decomposed into the sum of the two mutual information $I(\bz; \cdot)$ + $I(\bs ; \cdot)$. Thus, similarly to InfoGAN (see Eq. \ref{eq:InfoGAN-I}), we can retrieve four lower bounds.
As in \cite{chen_infogan_2016,lin_infogan-cr_2020}, to promote stability and efficiency, we model the two auxiliary distributions, $Q_z$ and $Q_s$, as factorized distributions.
Beside a factorized Gaussian distribution with identity covariance, we have also tested a factorized Laplace distribution $\bL(\mu,b)$ with $b=1$. This brings to a $l1$ reconstruction loss instead of a standard $l2$, and showed better performance in practice.\\
Finally, to better train $Q_s$, and since we know that $\bs$ should be equal to 0 for real images of domain $X$ (i.e., $\bx_R \sim P(\bx_R)$), we also add as regularization the lower bound of the mutual information $I(\bs;\bx_R)$. 
As before, we fix $P_x(\bs) = \delta(\bs = 0)$. The sum of these five lower bounds defines the $\mathcal{L}_{Info}$ loss: 
\begin{equation}
\begin{split}
    &\mathcal{L}_{Info}(G,Q_z,Q_s) = w_{bg} \mathbb{E}_{\bz \sim P_y(\bz)}\bigl[w_{Info}^z |(Q_z(G(z,0))- z|\\
    &+w_{Info}^s |Q_s(G(z,0))- 0|\bigr] \\ 
    &+w_t \mathbb{E}_{\bz,\bs \sim P_y(\bz,\bs)}\bigl[w_{Info}^z  |(Q_z(G(z,s))- z|\\
    &+ w_{Info}^s |Q_s(G(z,s))- s|\bigr] \\
    &+w_{Info}^{real} \mathbb{E}_{\bx_R \sim P(\bx_R)}\bigl[|(Q_s(\bx_R))- 0|\bigr]
\end{split}
\end{equation}
\noindent \textbf{Image reconstruction loss} \quad Differently from usual GAN models, we also propose to maximize the log-likelihood $\log(P(\by))$ (and $\log(P(\bx))$) of the generated images based on the proposed model. Indeed, no likelihood is generally available for optimizing the generator $G$ in a GAN model \cite{Goodfellow2014}. However, here, given a real image $\by_R$ (or $\bx_R$), we can use the auxiliary encoder $Q=(Q_s,Q_z)$ to estimate the latent factors $\hat{z}$ and $\hat{s}$ that should generate $\by_R$ (or $\bx_R$) and then maximize (an approximation) of the log-likelihood of the generated images $\by=G(\hat{z},\hat{s})$ (or $\bx=G(\hat{z},0)$):
\begin{equation}
\begin{split}
 \log P(\by) &\geq \bbE_{\by_R \sim P(\by_R), (\bz,\bs) \sim Q(\bz, \bs |\by_R)} \log P(\by|\bz,\bs,\by_R) \\&- \bbE_{\by_R \sim P(\by_R)} KL(Q(\bz,\bs|\by_R) || P(\bz,\bs|\by_R))
\end{split}
\end{equation}
We notice that the second term should tend towards 0 during training thanks to the previous Info Loss.\footnote{Lower bounds become tight as $Q$ resembles the true $P$.} We can thus approximate $\log P(\by)$ by computing only the left term and modeling $P(\by|\bz,\bs,\by_R)$ as a Laplace distribution $\bL(\mu,b)$ with 
$b=1$. We use a Laplace distribution, instead of a Gaussian one, since it has been shown, for instance in \cite{isola_image--image_2017}, that a $l1$-loss encourages sharper and better image reconstructions than a $l2$-loss. Similar computations can be done for $\log P(\bx)$. We define $\mathcal{L}_{Im}(G,Q_z,Q_s)= \log P(\bx) + \log P(\by)$:
\begin{equation}
\begin{split}
    \mathcal{L}_{Im}(G,Q_z,Q_s) &= w_{bg} \mathbb{E}_{ \bx_R \sim P(\bx_R), \hat{z}=Q_z(\bx_R)}\bigl[|G(\hat{z},0) - \bx_R|\bigr] \\
    &+  w_t \mathbb{E}_{\by_R \sim P(\by_R), \hat{z}, \hat{s}=Q(\by_R)}\bigl[|G(\hat{z},\hat{s}) - \by_R|\bigr]
 \end{split}
\end{equation}

\section{Results}\label{results}
In this section, we present the results of our model on four different visual datasets. Three of them  (CelebA with accessories \cite{weinberger_moment_2022}, Cifar-10-MNIST and dSprites-MNIST) have been conceived for the CA setting, giving us the possibility to qualitatively and quantitatively evaluate the performance of our model. We compare it with two SOTA Contrastive VAE algorithms (cVAE \cite{abid_contrastive_2019} and MM-cVAE  \cite{weinberger_moment_2022}) that had the best results in \cite{weinberger_moment_2022}.\footnote{We use the code provided by the authors of MM-cVAE.} The fourth dataset, Brats \cite{menze2014multimodal}, comprises T1-w MR brain images of healthy subject and patient with brain tumors, and is used for qualitative evaluation.\\ 
For quantitative evaluation, we use the fact that the information about attributes (e.g. glasses/hats in CelebA, MNIST digits, Cifar objects) should be present either in the common or in the salient space. Given a test set of images, we first use $Q$ to reconstruct $\hat{z}$ and $\hat{s}$ and then train a classifier on them to predict the attribute presence. By evaluating the discriminative power of the classifier, we can understand whether the information about the attributes has been put in the common or salient latent space by the method.\\
Qualitatively, the model can be evaluated by: 1) looking at the image reconstruction, 2) generating new images (sampling different salient features) and 3) swapping salient features. Given two real images $\bx_R \in X$ and $\by_R \in Y$, we can first estimate the latent factors $(\hat{z_X}, \hat{s_X})$ and ($\hat{z_y},\hat{s_Y})$, that should have generated $\bx_R$ and $\by_R$, using $Q$. Then, we can swap the estimated salient features $\hat{s_X}$ and $\hat{s_Y}$, and re-generate the images $G(\hat{z_X},\hat{s_Y})$ and $G(\hat{z_Y},\hat{s_X})$.\\ 
Implementation details about the architectures and hyper-parameters used in the different experiments can be found in the Supplementary.
\begin{table}
\centering
\hskip-0.33cm
\begin{adjustbox}{width=0.5\textwidth}
\setlength{\tabcolsep}{5pt} 
\renewcommand{\arraystretch}{0.9} 
\begin{tabular}{ |c|c|c|c|c|c|c| } 
 \hline
         & \multicolumn{3}{|c|}{$\hat{s}_y$ $\uparrow$} & \multicolumn{3}{|c|}{$\hat{z}_y$ $\downarrow$} \\ 
         & Best & Average & Worst & Best & Average & Worst\\
\hline
 cVAE*       & $0.84$ & $0.82$ & $0.81$ & $0.78$ & $0.80$ & $0.81$ \\ 
 MM-cVAE* & $0.85$ & $0.82$ & $nan$& $0.72$ & $0.76$ & $nan$ \\ 
 double InfoGAN &  $\mathbf{0.95}$ & $\mathbf{0.95}$ & $\mathbf{0.94}$ & $\mathbf{0.69}$ & $\mathbf{0.71}$ & $\mathbf{0.73}$ \\
 \hline
\end{tabular}
\end{adjustbox}
\caption{5-fold average accuracy on Target CelebA (glasses vs hat). Std is always $\leq$ 0.01, so we don't report it for clarity. Best results in \textbf{bold}.\\ \scriptsize{*: Results are different from  \cite{weinberger_moment_2022} where no external test set is used.}}
\label{tab:target-sep-celeba}
\end{table}

\paragraph{CelebA with accessories}  We use the dataset based on CelebA~\cite{liu2015faceattributes} presented in \cite{weinberger_moment_2022}, where background images $X$ are faces with neither hat of glasses, and target images $Y$ are faces with hat or glasses.
We use 20,000 images for training, 10,000 background and 10,000 target, equally divided between glasses and hat.  To evaluate the target class separation, we create a test set with images (5,000 with glasses and 5,000 with hat) never seen during training and compute the accuracy of a logistic regression (with 5-fold cross validation) on the reconstructed latent factors $\hat{s}_y=Q_s(\bs|\by)$ and $\hat{z}_y=Q_z(\bz|\by)$. Results are available in Table~\ref{tab:target-sep-celeba}. Please note that the evaluation protocol in \cite{weinberger_moment_2022} was different since authors did not use an external test set.
\noindent For a fair comparison, we run all methods 5 times (with different random seeds) for 500 epochs, and reported the highest (best), average and lowest (worst) scores. Extensive results are presented in the Supplementary. It is interesting to underline that MM-cVAE~\cite{weinberger_moment_2022} does not converge at every run. We have observed a divergence of the KL loss in about 10\% of the trainings,  which led to a convergence failure. We have used the original architecture of the MM-cVAE paper~\cite{weinberger_moment_2022} to reproduce their results.\\
We provide qualitative results in Fig.~\ref{fig:img-swap-celeba} with image reconstruction and salient feature swap. Please note that this would not be possible with SOTA IMI methods, such as CycleGAN~\cite{zhu_unpaired_2017} and MUNIT~\cite{huang_multimodal_2018}, not conceived for the CA setting. 
First of all, we observe that our model produces images of better quality than MM-cVAE, although this could probably be improved using larger GAN architectures, such as BigGAN \cite{Brock2019} or StyleGAN \cite{Karras2019}. From a quantitative point of view, our model obtains an average Inception Score (IS) equal to $1.63 \pm 0.03$ for background images and $2.66 \pm 0.02$ for target images, whereas MM-cVAE obtains $1.43 \pm 0.03$ and $1.44 \pm 0.01$ for background and target images respectively. Similar results were obtained using the Fréchet inception distance (FID).\\
It is interesting to notice that our model, contrarily to MM-cVAE, preserves the characteristics of the salient elements, such as the opacity of the glasses. Both models struggle to preserve the shape of the original hat, although our method tends to generate a better hat but based on the hairstyle of the person.\\ 
In Fig.~\ref{fig:fake-img-celeba}, we present qualitative results where we generate images fixing a $\bz$ in each row and using different $\bs$ (0 for $X$, $\neq 0$ for $Y$). 
We can see that there is indeed a change of domain, and that the model generates a wide variety of images. When switching from background $X$ to target $Y$, the characteristics of the person are well preserved, and a salient feature is added, here glasses or hat.
Furthermore, we can also notice that our model, being more accurate, is also more sensitive to dataset biases. For instance, we have noticed that in our dataset people with thin, transparent glasses are usually old men. This bias is clearly visible in the second row of Fig.\ref{fig:img-swap-celeba} and Fig.\ref{fig:fake-img-celeba}. Removing such bias, as in \cite{barbano_unbiased_2023}, is left as future work.

\begin{table}
\centering
\begin{adjustbox}{max width=.46\textwidth}
\setlength{\tabcolsep}{1pt} 
\renewcommand{\arraystretch}{0.3} 
\begin{tabular}{c|c c | c c }
    \textbf{Original} &  \multicolumn{2}{c}{\textbf{Reconstruction}}  & \multicolumn{2}{c}{\textbf{Swap}} \\
                        & MM-  & double & MM- & double  \\
                        & cVAE &  InfoGAN & cVAE & InfoGAN \\
    \hline\\
    \includegraphics[width=1.7cm]{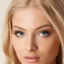} & \includegraphics[width=1.7cm]{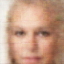} & \includegraphics[width=1.7cm]{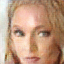} & \includegraphics[width=1.7cm]{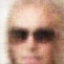} & \includegraphics[width=1.7cm]{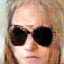} \\
     \includegraphics[width=1.7cm]{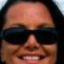} & \includegraphics[width=1.7cm]{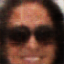} & \includegraphics[width=1.7cm]{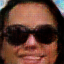} & \includegraphics[width=1.7cm]{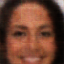} & \includegraphics[width=1.7cm]{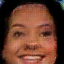} \\
    \hline \\
        \includegraphics[width=1.7cm]{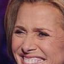} & \includegraphics[width=1.7cm]{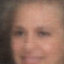} & \includegraphics[width=1.7cm]{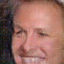} & \includegraphics[width=1.7cm]{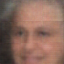} & \includegraphics[width=1.7cm]{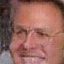} \\
         \includegraphics[width=1.7cm]{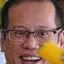} & \includegraphics[width=1.7cm]{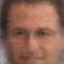} & \includegraphics[width=1.7cm]{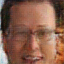} & \includegraphics[width=1.7cm]{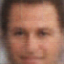} & \includegraphics[width=1.7cm]{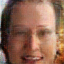} \\
        \hline \\
        \includegraphics[width=1.7cm]{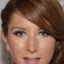} & \includegraphics[width=1.7cm]{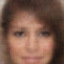} & \includegraphics[width=1.7cm]{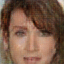} & \includegraphics[width=1.7cm]{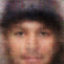} & \includegraphics[width=1.7cm]{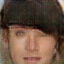} \\
         \includegraphics[width=1.7cm]{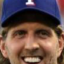} & \includegraphics[width=1.7cm]{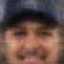} & \includegraphics[width=1.7cm]{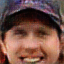} & \includegraphics[width=1.7cm]{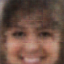} & \includegraphics[width=1.7cm]{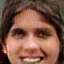} \\
    \end{tabular}
    \end{adjustbox}
    \captionof{figure}{Image reconstruction and swap with the CelebA with accessories dataset.}
    \label{fig:img-swap-celeba}
\end{table}


\begin{table}
\centering
\begin{adjustbox}{max width=.42\textwidth}
\setlength{\tabcolsep}{1pt} 
\renewcommand{\arraystretch}{0.2} 
    \begin{tabular}{c|c c c}
    \textbf{$X$}  &  \multicolumn{3}{c}{\textbf{$Y$ }} \\

    $G(\bz,0)$ & \multicolumn{3}{c}{$G(\bz,\bs^i\neq 0)$} \\
    
         \includegraphics[width=1.7cm]{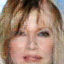} & \includegraphics[width=1.7cm]{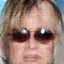} & \includegraphics[width=1.7cm]{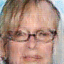} & \includegraphics[width=1.7cm]{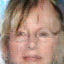}  \\

        \includegraphics[width=1.7cm]{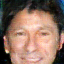} & \includegraphics[width=1.7cm]{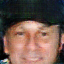} & \includegraphics[width=1.7cm]{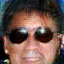} & \includegraphics[width=1.7cm]{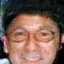}  \\

         \includegraphics[width=1.7cm]{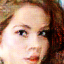} & \includegraphics[width=1.7cm]{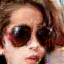} & \includegraphics[width=1.7cm]{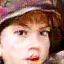} & \includegraphics[width=1.7cm]{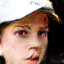}  \\

        \includegraphics[width=1.7cm]{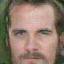} & \includegraphics[width=1.7cm]{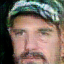} & \includegraphics[width=1.7cm]{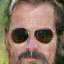} & \includegraphics[width=1.7cm]{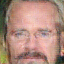}  \\
        
        \includegraphics[width=1.7cm]{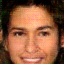} & \includegraphics[width=1.7cm]{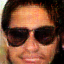} & \includegraphics[width=1.7cm]{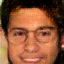} & \includegraphics[width=1.7cm]{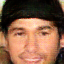}  \\
    \end{tabular}
    \end{adjustbox}
    \captionof{figure}{Fake images generated by our model. In each row, we use the same common feature $\bz$ for all images, $\bs=0$ for $X$ and different salient features $\bs \neq 0$ for $Y$.}
    \label{fig:fake-img-celeba}
\end{table}

\begin{table*}[h]
\centering
\begin{adjustbox}{width=0.88\textwidth}
\setlength{\tabcolsep}{3pt} 
\renewcommand{\arraystretch}{1} 
\begin{tabular}{|c|c|c|c|c|c|c|c|c|c|c|c|c|}
 \hline
\multicolumn{1}{|c|}{}     & \multicolumn{6}{c|}{Mnist (salient)}                                                                                                                                    & \multicolumn{6}{c|}{Cifar (background)}                                                                                                                                 \\
\multicolumn{1}{|c|}{}     & \multicolumn{3}{c|}{$\bs_y$ $\uparrow$}                                            & \multicolumn{3}{c|}{$\bz_y$ $\downarrow$}                                          & \multicolumn{3}{c|}{$\bs_y$ $\downarrow$}                                          & \multicolumn{3}{c|}{$\bz_y$ $\uparrow$}                                            \\
\multicolumn{1}{|c|}{}     & \multicolumn{1}{c|}{Best} & \multicolumn{1}{c|}{Avg.} & \multicolumn{1}{c|}{Worst} & \multicolumn{1}{c|}{Best} & \multicolumn{1}{c|}{Avg.} & \multicolumn{1}{c|}{Worst} & \multicolumn{1}{c|}{Best} & \multicolumn{1}{c|}{Avg.} & \multicolumn{1}{c|}{Worst} & \multicolumn{1}{c|}{Best} & \multicolumn{1}{c|}{Avg.} & \multicolumn{1}{c|}{Worst} \\ 
 \hline
MM-cVAE (size 128) & $0.81$  & $0.76$ & $nan$ & $0.43$ & $0.48$ & $nan$ & $0.14$ & $0.18$ & $nan$ & $0.36$ & $0.35$  & $nan$ \\
MM-cVAE (size 200) & $0.82$ & $0.63$ & $0.13$ & $0.43$ & $0.58$ & $0.82$ & $\mathbf{0.12}$ & $\mathbf{0.17}$ & $0.27$  & $0.37$ & $0.36$ & $0.34$  \\
double InfoGAN (size 128) & $0.87$ & $\mathbf{0.87}$ & $\mathbf{0.86}$  & $\mathbf{0.25}$  & $\mathbf{0.26}$  & $\mathbf{0.28}$  & $0.17$  & $0.18$ & $\mathbf{0.19}$  & $0.43$  & $0.42$  &  $0.41$  \\
double InfoGAN (size 200) & $\mathbf{0.88}$ & $\mathbf{0.87}$  & $\mathbf{0.86}$  & $0.32$  & $0.32$  & $0.32$  & $0.20$  & $0.21$ & $0.23$  & $\mathbf{0.44}$  & $\mathbf{0.44}$  & $\mathbf{0.43}$    \\
 \hline
\end{tabular}
\end{adjustbox}
\caption{
MNIST-Cifar10 classification. Digits information should only be encoded in $\bs_y$ and not in $\bz_y$, whereas the contrary should be true for Objects information. Std $\leq$ 0.01.
Best results in \textbf{bold}.}
\label{tab:cifar-mnist}
\end{table*}

\begin{table}
\centering
\begin{adjustbox}{max width=.43\textwidth}
\setlength{\tabcolsep}{3pt} 
\renewcommand{\arraystretch}{0.8} 
\begin{tabular}{|c |cc|cc|} 
\hline
 & \multicolumn{2}{c|}{Mnist (salient)} & \multicolumn{2}{c|}{Cifar (bg)}\\
 & $s_y$ $\uparrow$ & $z_y$ $\downarrow$  & $s_y$ $\downarrow$ & $z_y$ $\uparrow$ \\ 
 \hline
 -$L_{Class}$ & 0.48 & 0.83 & 0.23 & 0.37 \\												
- $L_{Class}$ - $L_{Im}$ & 0.54 & 0.72 & 0.22 & 0.38 \\	
- $L_{Info}$ - $L_{Class}$ - $L_{Im}$ & 0.70 &	0.70 &	 \textbf{0.18} & 0.18 \\						
- $L_{Info}$ & 0.85 & 0.60 & 0.30 & 0.36 \\				
- $L_{Info}$ - $L_{Im}$ & 0.59 & 0.59 & 0.20 & 0.20 \\
- $L_{Class}$ - $L_{Info}$	& 0.74 & 0.67 & 0.29 & 0.35 \\
- $L_{Im}$	& 0.86 & \textbf{0.25} &  0.20 &  \textbf{0.42} \\
Full & \textbf{0.87} & 0.26 &  \textbf{0.18} & \textbf{0.42} \\						
\hline
\end{tabular}
\end{adjustbox}
\caption{Ablation study of the different losses on the Cifar-MNIST dataset. For every configuration, 3 trainings were launched. We report average values.
}
\label{tab:ablation}
\end{table}

\paragraph{Cifar-10-MNIST dataset} We create a new dataset based on Cifar-10 \cite{krizhevsky_learning_2009} and MNIST \cite{lecun_mnist}. Background images $X$ are Cifar-10 images, and target images $Y$ are also CIFAR-10 with a random MNIST digit overlaid on it. We use 50k training images, equally divided between $X$ and $Y$, and 10k test images, equally divided among the MNIST digits. Our model should successfully capture the background variability (\textit{i.e.,} CIFAR objects) only in the common latent space $\bz_y$, and the MNIST variability (\textit{i.e.,} digits) only in the salient space $\bs_y$. 
A perfect classifier would have 100\% accuracy on MNIST when using $\bs_y$ and 10\% (which corresponds to randomness) when using $\bz_y$. Conversely, it should have 100\% accuracy on Cifar-10 when trained on $\bz_y$ and 10\% when trained on $\bs_y$.

We compare our model with MM-cVAE. Since we used the same image size as for CelebA ($64\times64\times3$), we kept the same network architecture. We tested several hyper-parameters for both methods and used the best configuration in our experiments. Results using two different latent space size are shown in Table \ref{tab:cifar-mnist} (for MM-cVAE, we use: $\lambda_1=10^2$, $\lambda_2=10^3$).  As before, we run both methods 5 times (with different random seeds) for 500 epochs, and report the highest, average and lowest scores. More results in the Suppl.\\
We can notice that our method either outperforms MM-cVAE or obtains comparable results. 
Moreover, during our numerous trainings, we noticed that the results obtained with our method are very stable, while those obtained with MM-cVAE, as before with CelebA, are more variable and may diverge (\textit{i.e.,} $nan$). 
Visual examples are presented in Fig.~\ref{fig:cifar-mnist-img}, with image reconstruction and salient feature swap (more in Supplementary). Our model offers sharper images than MM-cVAE and is able to better extract salient features.\\ 
\textbf{Ablation study} \quad 
We present in Table \ref{tab:ablation} a detailed ablation study on the proposed losses using the Cifar-MNIST dataset and the architecture with a latent space of size 128 (since it obtained the best results in Table \ref{tab:cifar-mnist}). We can notice that the proposed combination of losses obtains the best results.

\begin{adjustbox}{max width=.42\textwidth}
\setlength{\tabcolsep}{1pt} 
\renewcommand{\arraystretch}{0.2} 
\hskip 0.5cm
\begin{tabular}{c|c c | c c }
    \textbf{Original} &  \multicolumn{2}{c}{\textbf{Reconstruction}}  & \multicolumn{2}{c}{\textbf{Swap}} \\
                        & MM-  & double & MM- & double  \\
                        & cVAE &  InfoGAN & cVAE & InfoGAN \\
    \hline\\
       \includegraphics[width=1.7cm]{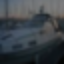} & \includegraphics[width=1.7cm]{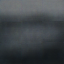} & \includegraphics[width=1.7cm]{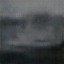} & \includegraphics[width=1.7cm]{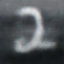} & \includegraphics[width=1.7cm]{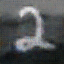} \\
       \includegraphics[width=1.7cm]{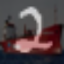} & \includegraphics[width=1.7cm]{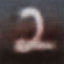} & \includegraphics[width=1.7cm]{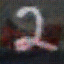} & \includegraphics[width=1.7cm]{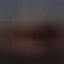} & \includegraphics[width=1.7cm]{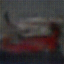}\\
        \hline\\
     \includegraphics[width=1.7cm]{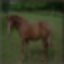} & \includegraphics[width=1.7cm]{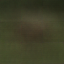} & \includegraphics[width=1.7cm]{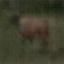} & \includegraphics[width=1.7cm]{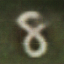} & \includegraphics[width=1.7cm]{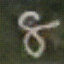} \\
     \includegraphics[width=1.7cm]{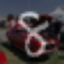} & \includegraphics[width=1.7cm]{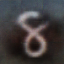} & \includegraphics[width=1.7cm]{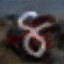} & \includegraphics[width=1.7cm]{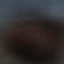} & \includegraphics[width=1.7cm]{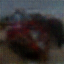} \\
        \hline \\
      \includegraphics[width=1.7cm]{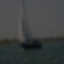} & \includegraphics[width=1.7cm]{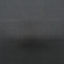} & \includegraphics[width=1.7cm]{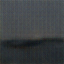} & \includegraphics[width=1.7cm]{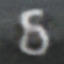} & \includegraphics[width=1.7cm]{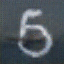} \\
     \includegraphics[width=1.7cm]{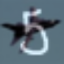} & \includegraphics[width=1.7cm]{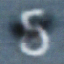} & \includegraphics[width=1.7cm]{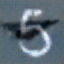} & \includegraphics[width=1.7cm]{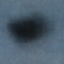} & \includegraphics[width=1.7cm]{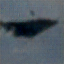} 
    \end{tabular}
    \end{adjustbox}
    \captionof{figure}{Image reconstruction and swap with Cifar-10-MNIST.}
    \label{fig:cifar-mnist-img}

\paragraph{Brats dataset}
In this section, we present qualitative results on the Brats dataset \cite{menze2014multimodal}. Background data $X$ contains T1-w MR brain images of healthy subject whereas the target dataset $Y$ has images of patients with brain tumors. 
Since images are bigger ($128 \times 128$) than the other datatsets, we use a different architecture. More details can be found in the Supplementary. Please note that here there are no sub-categories (as in previous datasets) that can be exploited to compute quantitative metrics (subgroup classification).\\
 Fig.~\ref{fig:fake-img-brats} shows fake images generated by our model trained on Brats. On the left are healthy images ($\bs=0$), and on the right images with tumor ($\bs \neq 0$). Images in the same row are generated using the same $\bz$. We can see that the general anatomy of the brain is preserved when changing domain, and that tumors with different size and position are generated. By changing $\bz$ (i.e. row), we can also notice that the model seems to have correctly encoded in $\bz$ the general anatomical variability of the brain.\\
 In Fig.\ref{fig:brats-counter}, we generate healthy counterparts of target images with tumor, setting $\bs=0$. This is very valuable in a clinical setting for multi-modal fusion \cite{francois_weighted_2022,maillard_deep_2022}, where images from different modalities can exhibit a different topology due to the tumor, and atlas construction \cite{liu_low-rank_2015,roux_mri_2019}, where tumor images have to be registered to healthy templates. Please note that here we use 2D slices with a small architecture (DCGAN), and a small (and biased) dataset (Brats). Indeed, we have noticed that most of the slices containing a tumor are in the central part of the brain (greater size) whereas slices from the higher or lower part of the brain (smaller size) have less frequently a tumor. This might thus entail structural changes during the generation of the healthy counterpart (swap), such as the one in size in the third row of Fig. 7. This could be solved by directly working with 3D data, more powerful networks and debiasing strategies.

\begin{table}
\centering
\begin{adjustbox}{max width=.42\textwidth}
    \setlength{\tabcolsep}{1pt} 
    \renewcommand{\arraystretch}{0.3} 
    \begin{tabular}{c|c c c}
    $X$ - \textbf{healthy}  &  \multicolumn{3}{c}{$Y$ - \textbf{tumor}} \\
    $G(z,0)$ & \multicolumn{3}{c}{$G(z,s^i\neq 0)$} \\

        \includegraphics[width=1.9cm]{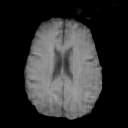} & \includegraphics[width=1.9cm]{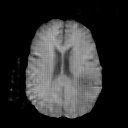} & \includegraphics[width=1.9cm]{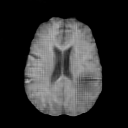} & \includegraphics[width=1.9cm]{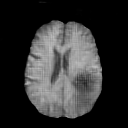}  \\

        \includegraphics[width=1.9cm]{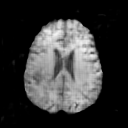} & \includegraphics[width=1.9cm]{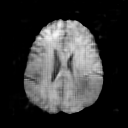} & \includegraphics[width=1.9cm]{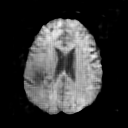} & \includegraphics[width=1.9cm]{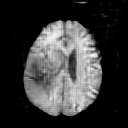}  \\

        \includegraphics[width=1.9cm]{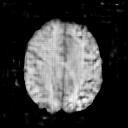} & \includegraphics[width=1.9cm]{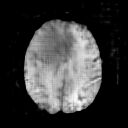} & \includegraphics[width=1.9cm]{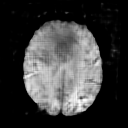} & \includegraphics[width=1.9cm]{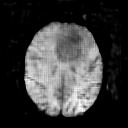}  \\
\end{tabular}
\end{adjustbox}
\captionof{figure}{Fake images generated by our model. In each row, we use the same $\bz$ for all images with $\bs=0$ for $X$ and different $\bs \neq 0$ for each exemple of $Y$.}
\label{fig:fake-img-brats}
\end{table}

\begin{table}
\centering
\begin{adjustbox}{width=.38\textwidth}
\setlength{\tabcolsep}{1pt} 
\renewcommand{\arraystretch}{0.2} 
     \begin{tabular}{c|c|c}
     \textbf{Target image} & \textbf{Reconstruction}  & \textbf{Swap ($\bs=0$)} \\      
     \includegraphics[width=2.5cm]{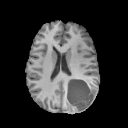} & \includegraphics[width=2.5cm]{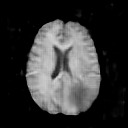} & \includegraphics[width=2.5cm]{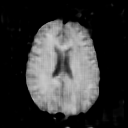} \\
     \includegraphics[width=2.5cm]{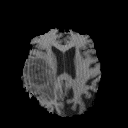} & \includegraphics[width=2.5cm]{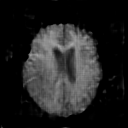} & \includegraphics[width=2.5cm]{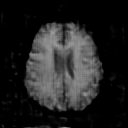} \\
     \includegraphics[width=2.5cm]{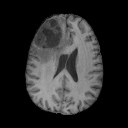} & \includegraphics[width=2.5cm]{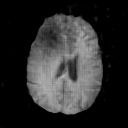} & \includegraphics[width=2.5cm]{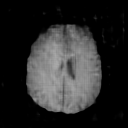} \\
     \end{tabular}
\end{adjustbox}
     \captionof{figure}{Reconstruction (middle) and generation of an healthy counterpart (swap, on the right) of a target image with brain tumor (on the left) by setting $\bs=0$ and keeping the same $\bz$.}
     \label{fig:brats-counter}
\end{table}

\paragraph{dSprites-MNIST dataset} A new toy dataset is proposed for evaluating CA methods. The background dataset $X$ consists of 4 MNIST digits (1, 2, 3 and 4) regularly placed in a square. In the target dataset $Y$, dSprites element \cite{dsprites17} are added on top of the same 4 MNIST digits.
Image reconstruction and salient feature swap are presented in Fig.~\ref{fig:dpsrite-img}. As before, we can see that, compared to MM-cVAE, image reconstructions are more accurate and sharp and, when exchanging salient features, the dSprites elements are better preserved.\\
\textbf{Disentanglement} As in \cite{higgins_-vae_2017,lin_infogan-cr_2020}, we also use dSprites to evaluate the disentanglement of our method in the salient space. Indeed, dSprites elements only exhibit 5 possible variations, making it easy to evaluate the disentanglement. Possible variations are: 1) shape (heart, elipse and square), 2) size, 3) position in X, 4) position in Y and 5) orientation (i.e. rotation). As metric, we use the FactorVAE (fvae) score \cite{kim_disentangling_2018}. Initial results using the proposed method showed a very poor disentanglement. 
To further improve it, we adapted for our model the Contrastive Regularizer (CR) module of InfoGAN-CR \cite{lin_infogan-cr_2020} (more details in the Supplementary), obtaining a maximum fvae score of 0.47. For comparison, InfoGAN-CR achieves a fvae score of 0.88 on the dsprite dataset alone. This shows that disentangling salient (or common) factors is much more difficult in our case than when using a single data-set. Exploring disentanglement regularizations more suited for a CA setting is left as future work.\\
In Fig.~\ref{fig:linspace}, we show target images generated by our model when varying only one dimension (from -1.5 to 1.5) of $\bs_y$, while keeping $\bz_y$ fixed. We clearly see a high entanglement among the dSprites factors of variation.
For completeness, we also checked whether the CR module helped the separation between common and salient information, and found similar quantitative results (see Supplementary). 
\section{Conclusions and Perspectives}
We propose the first GAN-based model for Contrastive Analysis (CA) that
estimates and separates in an unsupervised way all common and distinctive generative factors of a target dataset with respect to a background dataset. Compared to current SOTA CA-VAE models, we demonstrate superior performance on 4 visual datasets of increasing complexity and ranging from simple toy examples to real medical data. Our method manages to better separate common from salient factors, shows a better image generation quality and a greater stability during training. Furthermore, it allows the generation of multiple counterparts between domains by fixing the common factors and adding/removing the salient ones. We believe that the proposed method will benefit from more powerful GAN models and future progress in disentanglement, 
increasing its accuracy and interpretability. This will widen its fields of application to, for instance, clinically valuable and challenging tasks, such as computer aided-diagnosis. A last interesting research avenue could be the extension to the recent diffusion based models, as~\cite{song_score-based_2021,rombach_high-resolution_2022}.\\
\textbf{Limitations }
Recent works have shown that generative models, such as VAE and GAN, are in general not identifiable \cite{locatello_challenging_2019}. To obtain identifiability, two different solutions have been proposed: 1) either regularizing~\cite{kivva_identifiability_2022} / constraining (\textit{e.g.,} making it linear) the encoder or 2) introducing  an auxiliary variable so that the latent factors are conditionally independent given the auxiliary variable \cite{hyvarinen_nonlinear_2019,khemakhem_variational_2020}. Unfortunately, in Contrastive Analysis, neither of these solutions may be used\footnote{The dataset label could be considered as an auxiliary variable but it does not make $c$ and $s$ independent}. While all losses proposed here, and in the related works, are needed to effectively \textit{separate} common from salient factors, they do not assure that \textit{all} true generative factors have been identified. This is the main limitation of this work, and actually of all concurrent CA-VAE models, and is left as future work. Inspired by \cite{wyner_common_1975}, a possible research direction would be adding an information-theoretic loss that quantifies the common and salient information content so that, under realistic assumptions, the model could be identifiable.

\paragraph{Acknowledgments} This work was supported by the \textit{IDS department of Télécom Paris} and by the \textit{L’association Télécom Paris Alumni}.

\begin{table}
\centering     
\begin{adjustbox}{max width=0.4\textwidth}
\setlength{\tabcolsep}{2pt} 
\renewcommand{\arraystretch}{0.5} 
    \hspace{-0.4cm}
    \begin{tabular}{c|c c | c c }
    \textbf{Original} &  \multicolumn{2}{c}{\textbf{Reconstruction}}  & \multicolumn{2}{c}{\textbf{Swap}} \\
                        & MM-  & double & MM- & double  \\
                        & cVAE &  InfoGAN & cVAE & InfoGAN \\
        \includegraphics[width=1.7cm]{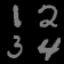} & \includegraphics[width=1.7cm]{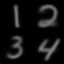} & \includegraphics[width=1.7cm]{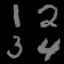} & \includegraphics[width=1.7cm]{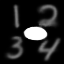} & \includegraphics[width=1.7cm]{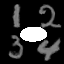} \\
        \includegraphics[width=1.7cm]{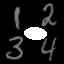} & \includegraphics[width=1.7cm]{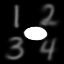} & \includegraphics[width=1.7cm]{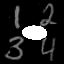} & \includegraphics[width=1.7cm]{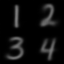} & \includegraphics[width=1.7cm]{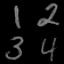} \\
        \hline \\
        \includegraphics[width=1.7cm]{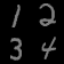} & \includegraphics[width=1.7cm]{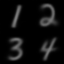} & \includegraphics[width=1.7cm]{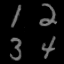} & \includegraphics[width=1.7cm]{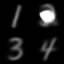} & \includegraphics[width=1.7cm]{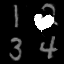} \\
        \includegraphics[width=1.7cm]{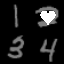} & \includegraphics[width=1.7cm]{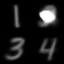} & \includegraphics[width=1.7cm]{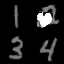} & \includegraphics[width=1.7cm]{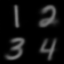} & \includegraphics[width=1.7cm]{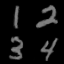} \\
    \end{tabular}
    \end{adjustbox}
    \captionof{figure}{Image reconstruction and swap of salient features on the dSprites-MNIST dataset.}
    \label{fig:dpsrite-img}
\end{table}

\begin{table}
\centering
\begin{adjustbox}{max width=.45\textwidth}
\setlength{\tabcolsep}{1pt} 
\renewcommand{\arraystretch}{0.2} 
\hspace{-0.4cm}
    \begin{tabular}{c c c c c c}
    \includegraphics[width=.2\textwidth]{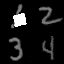} &
    \includegraphics[width=.2\textwidth]{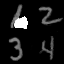} &
    \includegraphics[width=.2\textwidth]{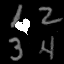} &
    \includegraphics[width=.2\textwidth]{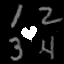} &
    \includegraphics[width=.2\textwidth]{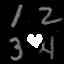} &
    \includegraphics[width=.2\textwidth]{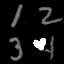} \\
    \includegraphics[width=.2\textwidth]{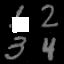} &
    \includegraphics[width=.2\textwidth]{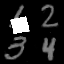} &
    \includegraphics[width=.2\textwidth]{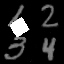} &
    \includegraphics[width=.2\textwidth]{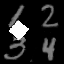} &
    \includegraphics[width=.2\textwidth]{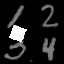} &
    \includegraphics[width=.2\textwidth]{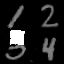} \\
    \includegraphics[width=.2\textwidth]{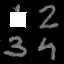} &
    \includegraphics[width=.2\textwidth]{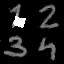} &
    \includegraphics[width=.2\textwidth]{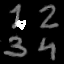} &
    \includegraphics[width=.2\textwidth]{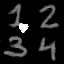} &
    \includegraphics[width=.2\textwidth]{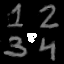} &
    \includegraphics[width=.2\textwidth]{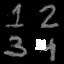} \\
    \includegraphics[width=.2\textwidth]{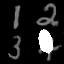} &
    \includegraphics[width=.2\textwidth]{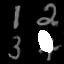} &
    \includegraphics[width=.2\textwidth]{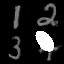} &
    \includegraphics[width=.2\textwidth]{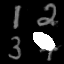} &
    \includegraphics[width=.2\textwidth]{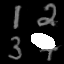} &
    \includegraphics[width=.2\textwidth]{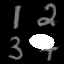}
\end{tabular}
\end{adjustbox}
\captionof{figure}{Each row represents the variation of only one element of the salient factor $\bs_y$, while keeping $\bz_y$ fixed. We can see a certain entanglement, with several parameters changing at the same time: shape and position (line 1), position and orientation (line 2).
Only the last line shows a disentanglement, with only the orientation of the ellipse changing.}
\label{fig:linspace}
\end{table}


\newpage
\clearpage
\newpage
\bibliography{DoubleInfoGAN}

\end{document}